\documentclass{article}

\usepackage{arxiv}

\usepackage[utf8]{inputenc} 
\usepackage[T1]{fontenc}    
\usepackage{hyperref}       
\usepackage{url}            
\usepackage{booktabs}       
\usepackage{amsfonts}       
\usepackage{nicefrac}       
\usepackage{microtype}      
\usepackage{lipsum}
\usepackage{graphicx}
\graphicspath{ {./figures/} }

\title{Deep Multimodal Fusion for Semantic Segmentation of Remote Sensing Earth Observation Data}

\author{
 Ivica Dimitrovski \\
  Faculty of Computer Science and Engineering\\
  University Ss Cyril and Methodius\\
  Skopje 1000, North Macedonia \\
  \texttt{ivica.dimitrovski@finki.ukim.mk} \\
   \And
 Vlatko Spasev \\
  Faculty of Computer Science and Engineering\\
  University Ss Cyril and Methodius\\
  Skopje 1000, North Macedonia \\
  \texttt{vlatko.spasev@finki.ukim.mk} \\
  \And
 Ivan Kitanovski \\
  Faculty of Computer Science and Engineering\\
  University Ss Cyril and Methodius\\
  Skopje 1000, North Macedonia \\
  \texttt{ivan.kitanovski@finki.ukim.mk} \\
}

\begin{document}
\maketitle
\begin{abstract}
Accurate semantic segmentation of remote sensing imagery is critical for various Earth observation applications, such as land cover mapping, urban planning, and environmental monitoring. However, individual data sources often present limitations for this task. Very High Resolution (VHR) aerial imagery provides rich spatial details but cannot capture temporal information about land cover changes. Conversely, Satellite Image Time Series (SITS) capture temporal dynamics, such as seasonal variations in vegetation, but with limited spatial resolution, making it difficult to distinguish fine-scale objects. This paper proposes a late fusion deep learning model (LF-DLM) for semantic segmentation that leverages the complementary strengths of both VHR aerial imagery and SITS. The proposed model consists of two independent deep learning branches. One branch integrates detailed textures from aerial imagery captured by UNetFormer with a Multi-Axis Vision Transformer (MaxViT) backbone. The other branch captures complex spatio-temporal dynamics from the Sentinel-2 satellite image time series using a U-Net with Temporal Attention Encoder (U-TAE). This approach leads to state-of-the-art results on the FLAIR dataset, a large-scale benchmark for land cover segmentation using multi-source optical imagery. The findings highlight the importance of multi-modality fusion in improving the accuracy and robustness of semantic segmentation in remote sensing applications.
\end{abstract}

\keywords{Earth observation, semantic segmentation, remote sensing, multi-modality fusion, deep learning}



\maketitle

\section{Introduction}\label{sec1}
Remote sensing data is captured from a distance by sensors or instruments mounted on various platforms such as satellites, aircraft, drones, and other vehicles. This data collects information about the Earth's surface, atmosphere, and other objects or phenomena without requiring direct physical contact \cite{toth2016remote}. There are two main techniques of remote sensing data acquisition: aerial and satellite. Satellite data is collected by satellites orbiting the Earth, capturing information over large areas at regular intervals. This provides a broad view of the entire planet. In contrast, aerial data is captured from airplanes or drones flying closer to the ground. This data covers smaller areas but with much finer detail, making it ideal for studying specific locations. Sensors are essential to remote sensing systems, as they collect data used to create images and other forms of information. Various types of sensors employed in remote sensing include optical sensors, radar sensors, lidar sensors, and electromagnetic sensors \cite{toth2016remote}.

Remote sensing data has four key properties: spectral, spatial, radiometric, and temporal resolution \cite{Spasev_2023}. Spectral resolution refers to the range of wavelengths a satellite sensor can detect. The more wavelengths a sensor can capture, the richer the information content of the imagery and the greater the detail it reveals about land use and cover. These captured wavelengths span a vast spectrum, including ultraviolet, visible light, near-infrared, infrared, and microwave. Some sensors capture just a few broad bands (multi-spectral), like Sentinel-2 with its 12 bands. Others, like Hyperion, are hyper-spectral, gathering thousands of narrow bands for a highly detailed spectral view \cite{Spasev_2023}. Spatial resolution refers to the size of each pixel in the image. Higher spatial resolution means smaller pixels, capturing finer details on the ground. Radiometric resolution describes how well the sensor can detect variations in radiated energy from the earth's surface. Higher resolution allows for better detection of subtle changes. Landsat 7 captures 8-bit images, distinguishing 256 distinct gray values of reflected energy, whereas Sentinel-2 features a 12-bit radiometric resolution, allowing it to discern 4095 gray values. Temporal resolution refers to how often a specific location is imaged. For example, polar-orbiting satellites exhibit varying temporal resolutions, ranging from 1 to 16 days (e.g., ten days for Sentinel-2). This is important for monitoring changes over time.

Machine learning is revolutionizing the way we analyze and understand remote sensing data \cite{DIMITROVSKI202318}. A particularly exciting area of research is the semantic segmentation of remote sensing data. The goal is to partition the image into meaningful regions, enabling detailed analysis and understanding of the Earth's surface. Accurate semantic segmentation of remote sensing imagery is essential for a wide range of Earth observation (EO) applications, including land cover mapping, urban planning, and environmental monitoring \cite{Spasev_2023}. The emergence of deep learning, particularly Convolutional Neural Networks (CNNs) and Fully Convolutional Networks (FCNs), ignited a revolution in semantic segmentation \cite{hao2020brief}. These models automated the learning of complex, hierarchical representations from data, paving the way for significant advancements. FCNs, often paired with encoder-decoder architectures, became the dominant approach. Early methods relied on successive convolutions and spatial pooling to generate dense predictions. Subsequent innovations like U-Net and SegNet introduced upsampling techniques to combine high-level features with lower-level ones during decoding \cite{hao2020brief}. This fusion aimed to capture both global context and precise object boundaries. To address the limited receptive field of standard convolutions in earlier layers of deep learning models, techniques like dilated (or atrous) convolutions were introduced by DeepLab \cite{chen2017deeplab}. These convolutions allow capturing a larger context while maintaining the resolution of the feature maps. Subsequent advancements incorporated spatial pyramid pooling (SPP) to capture multi-scale contextual information in higher layers, as seen in models like PSPNet \cite{hao2020brief} and UperNet \cite{xiao2018unified}. DeepLabV3+ built upon these advancements by combining atrous spatial pyramid pooling with a straightforward and efficient encoder-decoder architecture \cite{chen2018encoder}. However, recent developments like PSANet \cite{zhao2018psanet} and DRANet \cite{fu2020scene} have moved beyond traditional pooling, instead using attention mechanisms on top of encoder feature maps to capture long-range dependencies more effectively. Most recently, the adoption of transformer architectures, which utilize self-attention mechanisms and capture long-range dependencies, has marked additional advancement in semantic segmentation. Transformer encoder-decoder architectures like Segmenter, SegFormer, and MaskFormer harness transformers to enhance performance \cite{zheng2021rethinking}.

The abundance of diverse remote sensing modalities, like LiDARs, RGB-D cameras, and thermal cameras, has fostered the development of deep multimodal fusion techniques. These complementary sensors offer a richer picture of the scene, especially in complex environments. Deep learning excels at leveraging this data to reduce uncertainties and create a more comprehensive understanding. The core objective of deep multimodal fusion in segmentation is to learn an optimal joint representation by combining the strengths of individual modalities \cite{zhang2021deep}. This joint representation captures the rich and complementary features of the same scene, leading to more accurate segmentation results. For example, Very High Resolution (VHR) aerial imagery excels at providing rich spatial details, making it ideal for identifying fine-scale features such as individual buildings, roads, and small vegetation patches. This high level of detail is crucial for tasks that require precise mapping and analysis of specific locations. However, VHR aerial imagery typically lacks temporal information, which is essential for capturing changes over time to monitor dynamic processes such as seasonal variations in vegetation, urban growth, or the progression of environmental degradation. On the other hand, Satellite Image Time Series (SITS) data offers valuable temporal insights by capturing images of the same area at regular intervals. This capability is particularly useful for observing and analyzing temporal dynamics, such as the phenological cycles of crops, changes in land cover due to deforestation or reforestation, and the impact of natural disasters over time. However, SITS data generally has lower spatial resolution compared to VHR aerial imagery, which can make it difficult to distinguish fine-scale objects and detailed features on the ground. 

Deep multimodal fusion methods can be broadly categorized based on the stage at which information from different modalities is combined \cite{zhang2021deep}. Early fusion occurs at the raw data level (e.g., concatenating RGB and LiDAR data) or feature level (combining extracted features from each modality). This approach allows the model to learn a joint representation from the very beginning. Late fusion strategy involves processing each modality separately through individual deep learning branches. Then, the resulting feature maps are combined at a later stage (e.g., before the final prediction layer) using operations like concatenation, addition, or weighted voting. This approach offers greater flexibility in designing individual models for specific modalities. Hybrid fusion combines elements of both early and late fusion. It might involve initial feature-level fusion followed by late fusion of higher-level features. This allows for a more adaptive learning process based on the specific data and task. By effectively leveraging the complementary information from multiple modalities, deep multimodal fusion techniques are pushing the boundaries of semantic segmentation accuracy and robustness, particularly in complex remote sensing scenarios.

This paper tackles the challenge of accurate semantic segmentation in remote sensing by proposing a late fusion deep learning model (LF-DLM) that leverages the complementary strengths of VHR aerial imagery and SITS data. This approach aims to overcome the limitations inherent in single-source data, ultimately leading to more robust and informative land cover segmentation. The proposed LF-DLM architecture employs a dual-branch strategy, capitalizing on the specific advantages of each data source. Our research can be summarized by the following primary contributions:

\begin{itemize}
  \item Introduction of a late fusion deep learning model that leverages the complementary strengths of VHR aerial imagery and SITS data, tailored to enhance semantic segmentation of remote sensing imagery.
  \item Through comprehensive experimental evaluation, we demonstrate that the LF-DLM model effectively combines spatial and temporal information, leading to improved segmentation accuracy across various land cover types while maintaining efficient inference times.
  \item Our LF-DLM model achieves state-of-the-art results on the FLAIR dataset, surpassing previous benchmarks, thus establishing a new standard for semantic segmentation in multi-source optical imagery.
\end{itemize}

The subsequent sections of this paper are structured as follows: Section 2 provides an overview of the dataset utilized in the research. Section 3 details the key features of the proposed late fusion deep learning model. Section 4 comprehensively outlines the experimental design and setup, including data preprocessing, training protocols, model parameters, and evaluation metrics. Section 5 presents the experimental results alongside relevant discussions. Finally, Section 6 concludes the paper, summarizing the findings and contributions.

\section{Dataset}
The FLAIR dataset\footnote{\url{https://github.com/IGNF/FLAIR-2}} includes diverse sources of acquisition, each with unique characteristics and varying spatial, spectral, and temporal resolutions. This dataset provides detailed VHR aerial images, elevation models, and satellite image time series \cite{garioud2024flair}. Each aerial image measures 512~$\times$~512 pixels, with a spatial resolution of 20 cm per pixel, and includes four spectral bands: red, blue, green, and near-infrared. The dataset comprises 77762 patches. To ensure high-quality images, the aerial data is captured only during favorable weather conditions, specifically between April and November from 2018 to 2021. Each aerial image includes an elevation value. This value is derived from combining a digital elevation model and a digital surface model, obtained through photogrammetry on the aerial images, ensuring temporal consistency.

\begin{figure}[!ht]
    \centering
        \includegraphics[width=0.9\linewidth]{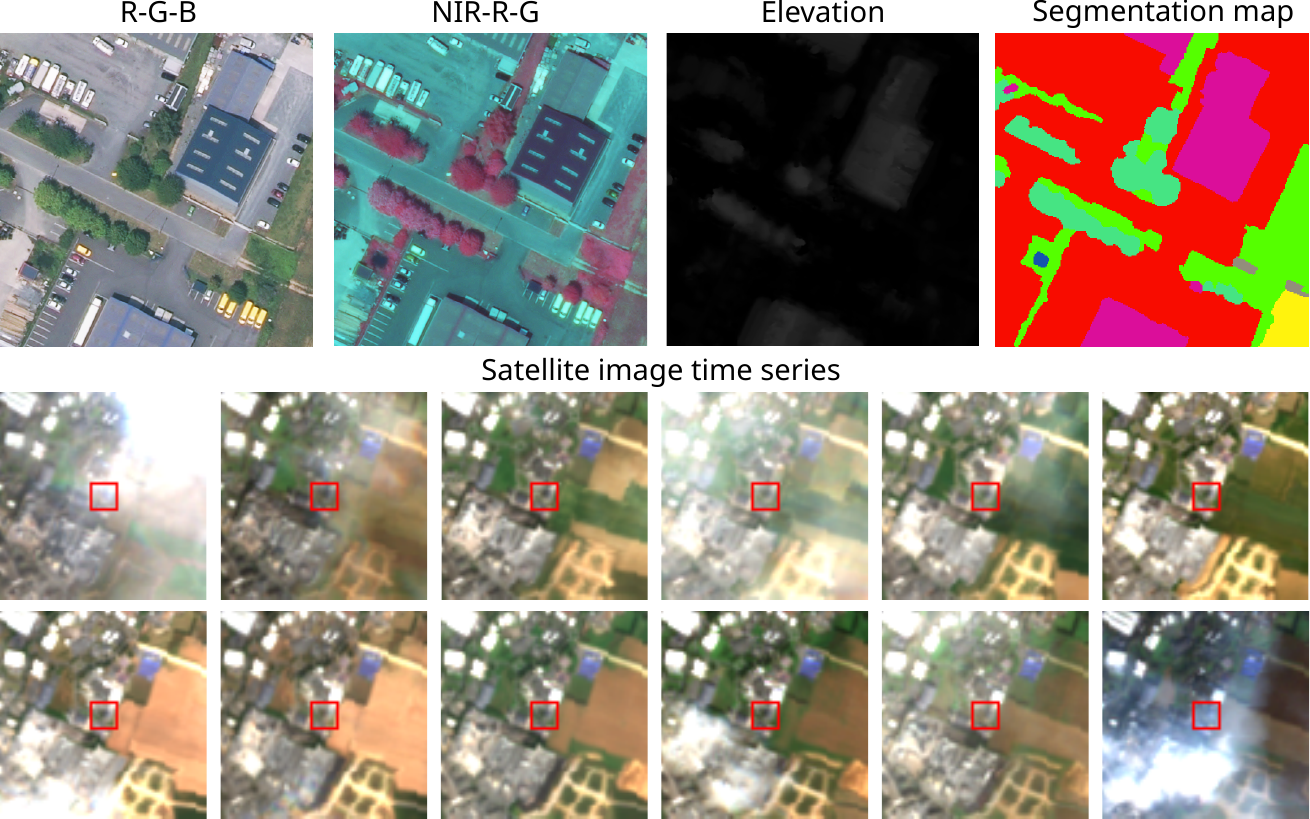}
    \caption{Example patches from the FLAIR dataset. Each patch contains an aerial image with red, green, blue (RGB), and near-infrared (NIR) values; a pixel-precise digital surface model providing an elevation for each pixel; segmentation map with labels for each pixel; and an optical time series from several months, centered on the aerial image. The red frame marks the area that corresponds to the aerial image.}
    \label{fig:dataset_samples}
\end{figure}

Each aerial image patch in FLAIR is accompanied by a corresponding time series of satellite images from the Sentinel-2 constellation \cite{DRUSCH201225}. These satellite images offer a broader view with a spatial resolution of 10 meters per pixel and come in a size of 40$\times$40 pixels, centered on the corresponding aerial image, and only 10$\times$10 center pixels correspond to the aerial image patch. Each pixel provides information across 10 spectral bands, capturing data from the visible to the medium infrared spectrum. The time series for each patch spans the entire year during which the aerial image was acquired. The number of images within a series can vary between 20 and 110, depending on satellite availability and orbital characteristics. The dataset includes acquisitions with cloud cover and provides cloud and snow probability masks, obtained with Sen2cor \cite{main2017sen2cor}, along with information about the satellite and its orbit. Example patches from the FLAIR dataset are given in Figure~\ref{fig:dataset_samples}.

\begin{figure}[!ht]
    \centering
        \includegraphics[width=0.6\linewidth]{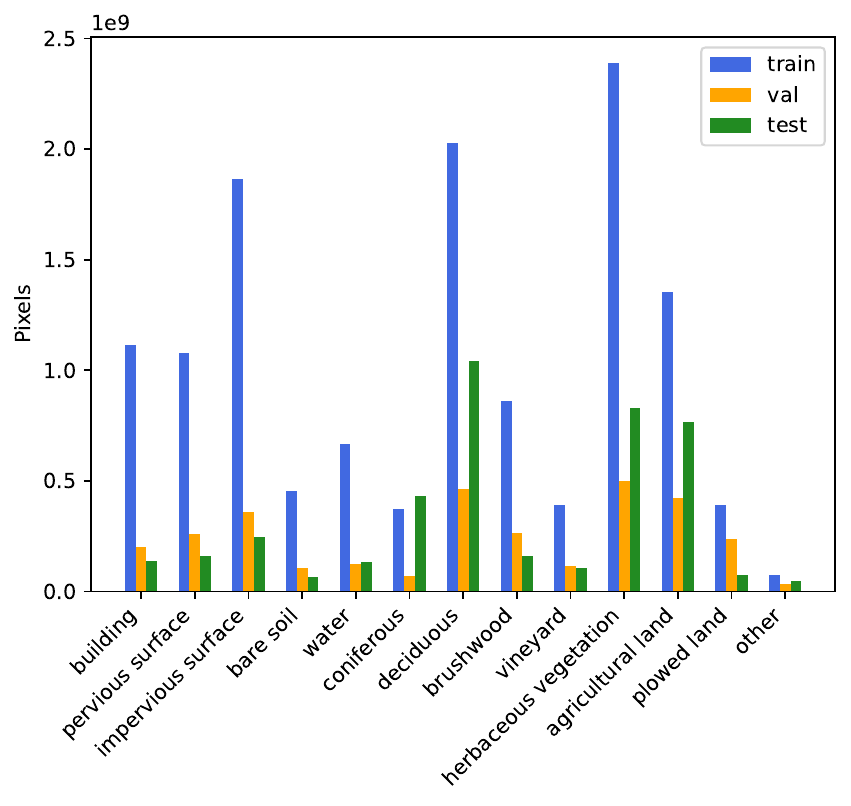}
    \caption{The distribution of pixels within the labels across the train, validation, and test sets of the FLAIR dataset.}
    \label{fig:dataset_distribution}
\end{figure}

The VHR images are annotated with segmentation masks containing 18 different labels/classes, along with an 'other' class for unknown land cover. Due to significant under-representation (less than 1\% of the complete dataset), five of these classes are combined into the 'other' class. This results in a nomenclature of 12 classes plus the 'other' class. The classes are: 'building', 'pervious surface', 'impervious surface', 'bare soil', 'water', 'coniferous', 'deciduous', 'brushwood', 'vineyard', 'herbaceous vegetation', 'agricultural land', and 'plowed land'. Annotations are not provided for the satellite images. Instead, these images are intended to support the aerial images by providing spatial context. The distribution of pixels within the labels across the train, validation, and test sets of the FLAIR dataset is shown in Figure~\ref{fig:dataset_distribution}.

The dataset comprises 50 spatial domains, each representing various landscapes and climates of metropolitan France. The training set includes 32 spatial domains, the validation set contains 8, and the remaining 10 domains are allocated to the test set. The dataset is part of the FLAIR \#2 challenge where a key requirement is to leverage both aerial and satellite imagery to achieve optimal semantic segmentation results. The FLAIR \#2 challenge introduces a second requirement: computational efficiency to ensure a balance between accuracy and practicality, considering the vast amount of data involved. The proposed approach's inference time needs to be within 2.5 times the execution speed of the baseline model offered by the dataset creators \cite{garioud2024flair}. Within this paper, we have carefully designed our solution to effectively utilize both data sources while staying within the allowed inference time for the FLAIR \#2 challenge.

\section{Model Architecture}
This paper introduces a late fusion deep learning model (LF-DLM) for semantic segmentation, designed to exploit the complementary strengths of both Very High Resolution (VHR) aerial imagery and Satellite Image Time Series (SITS). The proposed model features two independent deep learning branches. The first branch integrates detailed textures from aerial imagery using a UNetFormer \cite{wang2022unetformer} with a Multi-Axis Vision Transformer (MaxViT) encoder \cite{Tu_2022}, effectively capturing high-resolution spatial details. The second branch focuses on capturing complex spatio-temporal dynamics from the Sentinel-2 satellite image time series by employing a U-Net with Temporal Attention Encoder (U-TAE) \cite{garnot2021panoptic}, which processes and interprets temporal information. In the late fusion deep learning model, the probability scores from each branch are combined using a weighted geometric mean to obtain the final segmentation map. This dual-branch approach enables the model to leverage both spatial and temporal data for enhanced semantic segmentation performance.

The first branch of the LF-DLM model builds upon the Unet-like transformer UNetFormer. Originally it consists of a convolutional neural network (CNN) based encoder and a transformer-based decoder. The transformer-based decoder is constructed using global-local Transformer blocks (GLTB) that employ an efficient global-local attention mechanism with an attentional global branch and a convolutional local branch, enabling the capture of both global and local contexts for enhanced visual perception. We propose a modification to UNetFormer by replacing its CNN encoder with MaxViT, a hybrid vision transformer architecture. MaxViT introduces a novel building block called Multi-axis Self-Attention (Max-SA). This block allows the model to attend to information along multiple axes within an image feature map, including spatial, channel-wise axes, or combination. Compared to standard full self-attention in ViTs, Max-SA captures long-range dependencies (global information) more efficiently without requiring complex computations. MaxViT utilizes a hierarchical architecture where each stage in the hierarchy consists of a MaxViT block, which combines Max-SA with a convolutional layer. This combination leverages the strengths of both approaches: Max-SA for global context and convolutions for efficient local feature extraction. The network begins by downsampling the input through Conv3x3 layers in the stem stage (S0). The body of the network contains four stages (S1-S4), with each stage having half the resolution of the previous one with a doubled number of channels (hidden dimension). The feature maps generated by each stage are fused with the corresponding feature maps generated by the GLTB of the decoder using a weighted sum operation. 

The MaxViT model can be scaled up by increasing the number of blocks per stage and the channel dimension. There are several MaxViT variants including MaxViT-T, MaxViT-S, MaxViT-B, MaxViT-L, and MaxViT-XL. These variants progressively increase in complexity (number of blocks and channels) and likely performance, potentially reaching a trade-off between accuracy and efficiency \cite{Tu_2022}. In this study, we are using MaxViT-T as an encoder in the UNetFormer architecture. We are utilizing MaxViT-T, pre-trained on the ImageNet-1K dataset, to leverage its learned general visual features, which can be highly beneficial for semantic segmentation tasks.

To effectively analyze both spatial and temporal information within the Sentinel-2 satellite image time series, we leverage a U-Net with temporal attention (U-TAE) model, which serves as the second branch in the LF-DLM model. This branch extracts multi-scale spatio-temporal feature maps from SITS using a combination of spatial convolution and temporal attention. U-TAE encodes a given sequence in three key steps \cite{garnot2021panoptic}. First, each image in the sequence is embedded simultaneously and independently by a shared multi-level spatial convolutional encoder. Next, a temporal attention encoder collapses the temporal dimension of the resulting sequence of feature maps into a single map for each level. Finally, a spatial convolutional decoder produces a single feature map with the same resolution as the input images. By combining these steps, U-TAE allows effective exploitation of the rich spatio-temporal information present in the SITS, leading to a more comprehensive understanding of the scene dynamics.

\section{Experimental Design and Setup}
The primary objective of our study is to develop, evaluate, and compare a late fusion deep learning model (LF-DLM) for semantic segmentation of remote sensing imagery by leveraging the complementary strengths of VHR aerial imagery and SITS. Our experimental design is structured around the main hypothesis, that the fusion of these multi-source optical images will improve the semantic segmentation performance compared to using either data source alone. To test the hypothesis, we use VHR aerial imagery processed through the UNetFormer with MaxViT-S backbone to capture detailed spatial features, and SITS processed through a U-Net with Temporal Attention Encoder (U-TAE) to capture complex spatio-temporal dynamics. Our evaluation strategy involves training and assessing each model separately to determine their performances and conducting a comparative analysis to highlight the benefits of combining these data sources with weighted late fusion as a strategy.

The experimental setup involves data pre-processing, a configuration of the models, and hyperparameter selection. While no additional pre-processing is applied to the aerial patches, we address the potential influence of clouds and snow in the Sentinel-2 time series by implementing two pre-processing strategies using the provided mask files. Cloud filtering focuses on the probability of cloud or snow occurrence in the masks. We exclude images from the training process where the number of pixels exceeding a specific probability threshold (set to 0.5 in our experiments) surpasses a designated percentage of the total image pixels. This approach mitigates the impact of cloudy or snowy data on the training process. Additionally, we apply temporal monthly averaging to address challenges posed by the large number of dates within the time series. Here, a monthly average is computed using only cloudless dates within each month. If no cloudless dates are available for a particular month, the U-TAE branch might receive less than 12 images as input.

The UNetFormer model leverages a pre-trained MaxViT-T encoder on the ImageNet-1K dataset. This encoder receives five-channel aerial patches containing red, green, blue, near-infrared, and elevation data. The resolution of the aerial patches is $512\times512$ pixels. We add two channels to the initial layers to accommodate the near-infrared and elevation pixel values, with the weights of these added channels initialized randomly. The number of learnable parameters in this UNetFormer model is approximately 31 million. We use the default U-TAE parameters \cite{garnot2021panoptic}, \cite{garioud2024flair}, with the only modification being the widths of the encoder and decoder, which we adjusted to [64, 64, 128, 128]. This list specifies the number of channels for the successive layers of the convolutional encoder, and the same configuration applies to the decoder. The input to this model is the SITS data with dimensions $T\times10\times40\times40$, where $T$ represents the number of images in the time series (with a maximum of 12), 10 is the number of spectral bands, and $40\times40$ is the pixel resolution. The number of learnable parameters in this U-TAE model is approximately 2.9 million. To ensure spatial alignment, the U-TAE outputs are first cropped to match the size of the corresponding aerial patch. Then, they are upsampled to the same resolution ($512\times512$ pixels) as the aerial mask files. We experimented with different weight combinations for the late fusion, and the best performance was achieved when the UNetFormer branch was assigned a weight of 0.7 and the U-TAE branch had a weight of 0.3 in the weighted geometric mean.

To train both the UNetFormer and U-TAE models effectively, we employed several common deep learning techniques. To prevent overfitting and optimize hyperparameters, we employed a training process with hyperparameter selection and early stopping. The training data was used to train the model. Hyperparameter selection was performed on the validation split to identify the optimal configuration for the model's hyperparameters. To prevent overfitting, we implemented early stopping using the validation loss. If the validation loss did not improve for a predefined patience period (15 epochs in this case), training was terminated. The model with the best performance on the validation set, determined by the chosen evaluation metric, was then saved as the final model. This model was subsequently evaluated on the unseen test data to obtain an unbiased assessment of its predictive performance. The maximum training duration was set to 30 epochs. To improve the robustness and generalization ability of our model, we employ data augmentation techniques during training. This process involves applying random geometric transformations to the training data. Specifically, we utilize horizontal flips, vertical flips, and random rotations at predefined angles (0, 90, 180, and 270 degrees). We fixed the batch size at 12 for our experiments. We employed the AdamW optimizer with a learning rate of 0.0001 \cite{loshchilov2017decoupled}. A polynomial decay scheduler was used to gradually decrease the learning rate throughout training. This approach, with a carefully chosen decay rate, has been shown to improve model performance \cite{rs16122077}. The scheduler applies a polynomial function to the AdamW optimizer, starting with an initial learning rate of (1 $\times$ {10}${^{-4}}$) and reaching a final learning rate of (1 $\times$ {10}${^{-7}}$) within the specified number of decay steps. To achieve a balanced approach to semantic segmentation, we combined Cross Entropy Loss and Dice Loss. Cross Entropy Loss measures the similarity between predicted and ground truth masks at each pixel, while Dice Loss focuses on accurate boundary localization. This combination effectively addresses both object localization and overall segmentation accuracy.

All models were trained on NVIDIA A100-PCIe GPUs with 40 GB of memory running CUDA version 11.5. We configured and ran the experiments using the deep learning framework PyTorch Lightning \cite{Falcon_PyTorch_Lightning_2019}. In the experimental setup, we carefully considered the constraints and requirements of the FLAIR dataset, as it is part of the FLAIR \#2 challenge. Our solution was designed to effectively utilize both data sources while adhering to the allowed inference time for the FLAIR \#2 challenge. The FLAIR \#2 challenge specifies a maximum inference time that cannot exceed 2.5 times the baseline method. By measuring the inference time of the provided FLAIR \#2 challenge baseline code on our environment, we determined that it takes approximately 396 seconds to generate segmentation maps for all images in the test set.  Since the challenge restricts inference time to a maximum of 2.5 times the baseline, our model's inference time must not exceed 2.5 times 396 seconds, which translates to approximately 990 seconds. We assess the model performance using label-wise intersection over union ($IoU$) which denotes the area of the overlap between the ground truth and predicted label divided by the total area. We also report the mean intersection over union ($mIoU$) averaged across the different labels. The evaluation metrics are computed for the first 12 classes, excluding the 'other' class. 

\section{Results}
Table~\ref{tab:results_summary} summarizes the performance of each model on the FLAIR dataset. Label-wise Intersection over Union (IoU) and mean IoU (mIoU) are reported in percentage. As expected, the U-TAE model achieved the lowest mIoU (39.68\%) due to the limited spatial resolution of the satellite image time series. The UNetFormer model, leveraging the high spatial detail of aerial imagery, significantly improved upon this with a mIoU of 62.81\%, representing a 23.13\% increase. This outcome is expected given that the satellite imagery has a spatial resolution 50 times lower than the aerial imagery (10 m versus 0.2 m). Notably, the Late Fusion Deep Learning Model (LF-DLM) achieved the best overall performance with a mIoU of 63.10\%. This represents an improvement of 0.29\% compared to the UNetFormer alone. These findings support our hypothesis that combining information from both aerial imagery and satellite time series data through late fusion leads to improved semantic segmentation performance. 

\begin{table}[h] 
\centering
    \caption{Mean intersection over union (mIoU \%) and Intersection over Union (IoU \%) for each label of the UNetFormer, U-TAE, and LF-DLM models over the FLAIR dataset.}\label{tab:results_summary}
    \begin{tabular}{l|ccc}
    \toprule
        Label~\textbackslash~Model & UNetFormer & U-TAE& LF-DLM \\
        \midrule
        building & 85.40 & 35.53 & 85.14 \\
        pervious surface & 57.69 & 31.36 & 58.31 \\
        impervious surface & 74.95 & 38.67 & 74.66 \\
        bare soil & 63.89 & 39.02 & 65.01 \\
        water & 90.77 & 74.75 & 91.08 \\
        coniferous & 65.67 & 54.74 & 66.89 \\
        deciduous & 73.83 & 56.36 & 74.32 \\
        brushwood & 27.68 & 11.35 & 26.77 \\
        vineyard & 67.19 & 49.48 & 67.59 \\
        herbaceous vegetation & 50.93 & 26.80 & 50.86 \\
        agricultural land & 56.16 & 45.43 & 56.59 \\
        plowed land & 39.55 & 12.67 & 39.96 \\
        \midrule
        mIoU & 62.81 & 39.68 & 63.10 \\  
        \bottomrule
    \end{tabular}
\end{table}

The LF-DLM shows improvement in IoU for most labels compared to the UNetFormer model. This is evident in labels like 'pervious surface' (0.62\%), 'bare soil' (1.12\%), 'water' (0.32\%), 'coniferous' (1.22\%), 'deciduous' (0.5\%), 'vineyard' (0.41\%), 'agricultural land' (0.43\%), and 'plowed land' (0.41\%). This suggests that the late fusion strategy effectively combines the strengths of both U-TAE (capturing spatio-temporal information) and UNetFormer (capturing high spatial details) to improve segmentation accuracy across various land cover types. This is particularly evident for labels like 'coniferous' where SITS data, containing temporal information, might be crucial to distinguish them from 'deciduous' trees exhibiting seasonal changes in spectral properties. The improvement is also notable for the 'bare soil' label, potentially benefiting from the complementary information provided by SITS data. However, the LF-DLM shows a slight decrease in IoU for labels like 'building' (-0.26\%), 'impervious surface' (-0.29\%), 'brushwood' (-0.91\%), and 'herbaceous vegetation' (-0.07\%). This could be due to several factors like class/label imbalance or fusion complexity where the late fusion process might introduce additional complexity for these specific labels, leading to slight performance drops compared to the UNetFormer model. Potentially we can explore the possibility of employing label-specific weighting or fusion techniques during late fusion to potentially address challenges faced by specific land cover types. 

\begin{figure}[h]
  \includegraphics[trim={0 0 0 0},clip, width=0.7\linewidth]{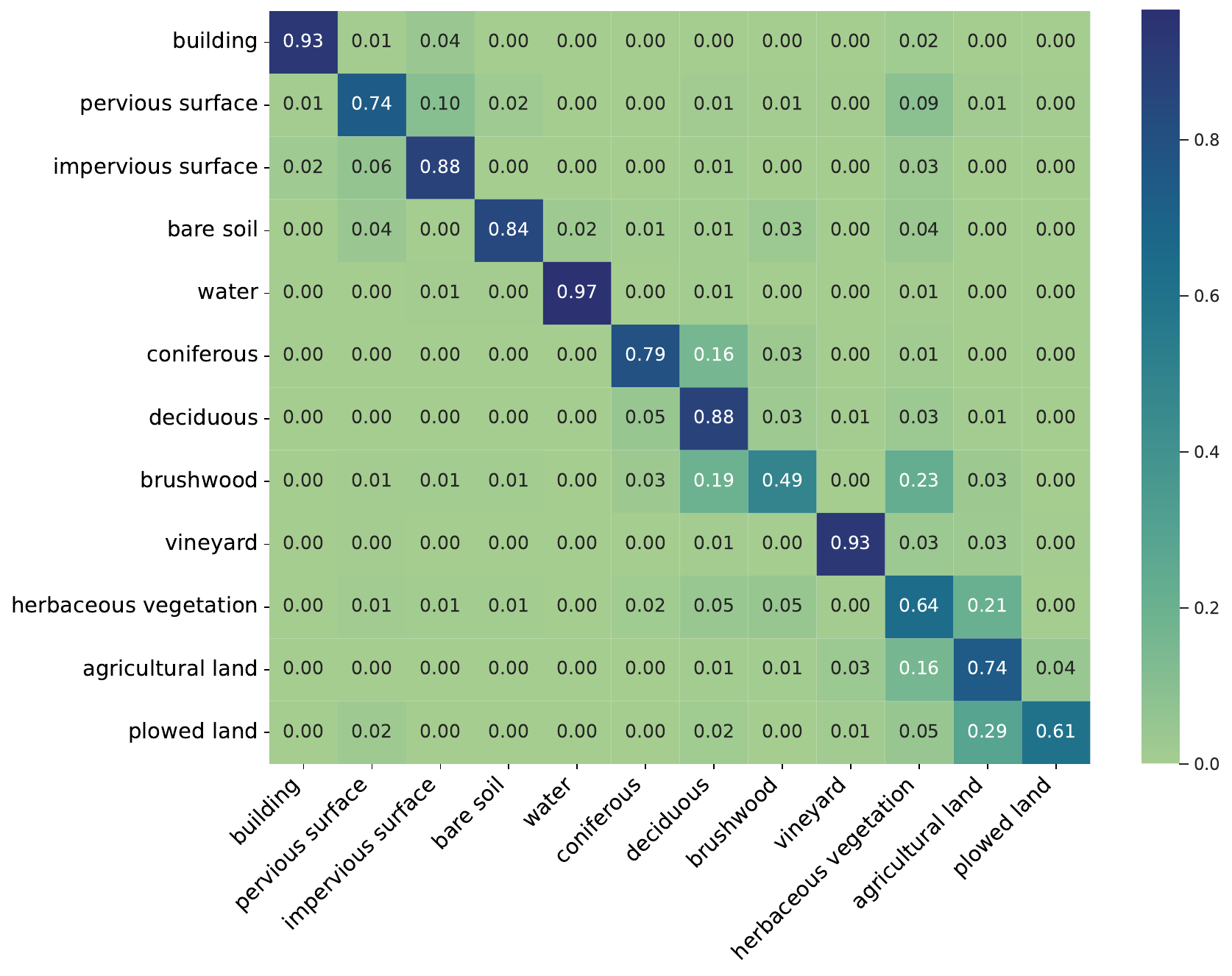}
  \caption{Confusion matrix for LF-DLM on the FLAIR~dataset.}
  \label{fig:flair_confmat}
\end{figure}

Examining the confusion matrix depicted in Figure~\ref{fig:flair_confmat} reveals that the best LF-DLM model achieves high prediction accuracy, with minimal misclassification in the majority of labels. However, it tends to confuse the labels "coniferous" and "deciduous", "brushwood" and "herbaceous vegetation", "brushwood" and "deciduous", as well as "agricultural land" and "herbaceous vegetation". This is rather expected given the semantic similarity between these labels. 

\begin{figure}[h]
        \includegraphics[width=0.9\linewidth]{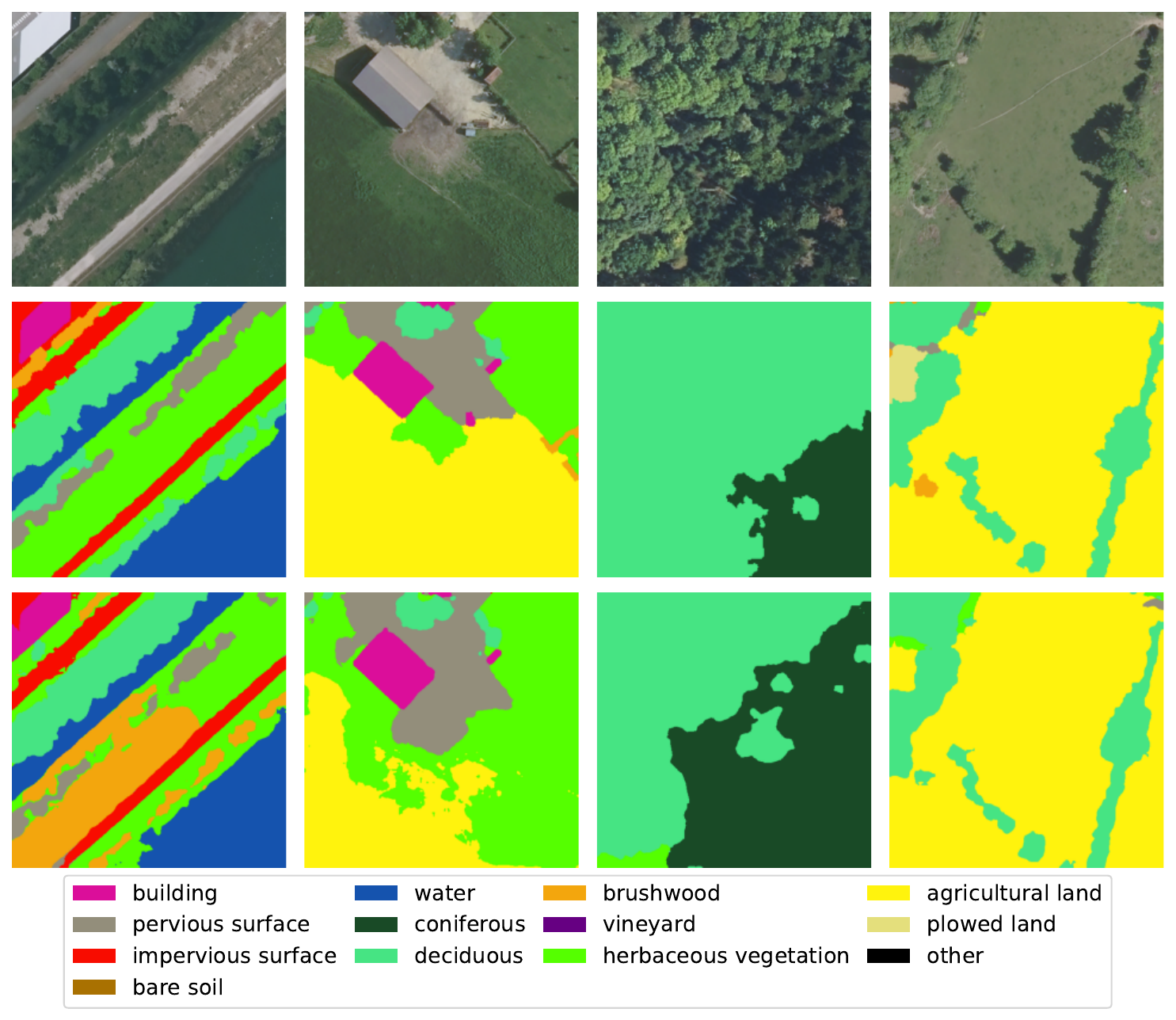}
    \caption{Example images, ground-truth masks, and inference masks from the FLAIR dataset. The first row shows example images. The second row shows the corresponding ground-truth masks. The third row shows the prediction results of the LF-DLM.}
    \label{fig:flair_inference}
\end{figure}

Figure~\ref{fig:flair_inference} shows several example images, ground truth masks, and predicted masks from the FLAIR dataset. Obtaining accurate segmentation maps is very challenging due to factors like complex scenes, occlusion between different land cover areas, and the semantic similarity between land cover types.

To ensure compliance with the inference time constraints of the FLAIR \#2 challenge, we measured the inference times of our proposed models. Table~\ref{tab:results_inference} presents these measurements along with their corresponding ratios compared to the baseline model inference time. All measurements were conducted on the same machine equipped with an NVIDIA A100-PCIe GPU with 40 GB of memory. The challenge restricts inference time to a maximum of 2.5 times that of the baseline model. As the table indicates, our models currently operate within this limit. This suggests potential for further improvement in our model's predictive performance while maintaining compliance with the challenge's constraints.

\begin{table}[h] 
\centering
    \caption{Inference times and their corresponding ratios compared to the baseline model inference time for our proposed models.}\label{tab:results_inference}
    \begin{tabular}{l|cc}
    \toprule
        Model & Inference time (sec.) & Relative time \\
        \midrule
        U-TAE & 229 & 0.58 \\
        UNetFormer & 429 & 1.08 \\
        LF-DLM & 594 & 1.5 \\
        \bottomrule
    \end{tabular}
\end{table}

To enhance predictive performance while adhering to the FLAIR \#2 challenge's inference time constraint, we incorporated a second UNetFormer model into the late fusion scheme. This second model was trained using identical parameters, with the sole difference being a variation in the random seed. The resulting late fusion deep learning model comprises the U-TAE model, two UNetFormer models (with different random seeds), and the late fusion layer. This configuration achieved a mIoU value of 64.52\%, an inference time of 943 seconds, and a relative inference time ratio of 2.38. Importantly, this remains well within the challenge's allowed constraints.

To comprehensively evaluate our best model configuration's predictive performance, we compared it with previously employed methods on the FLAIR dataset. The challenge organizers provided a U-Net baseline with a ResNet34 backbone in combination with a U-TAE model using a mid-stage fusion of features from both models \cite{garioud2024flair}, achieving a mIoU of 57.58\%. Our best model surpasses the baseline by a significant margin of 6.94\%. The current state-of-the-art on this dataset was an ensemble model consisting of four base models. The base models are similar to the baseline model provided by the challenge organizers, the only modification is the replacement of the ResNet34 backbone with MiT and ResNeXt backbones in the U-Net model \cite{straka2024modernized}. Additionally, a two-stage training procedure is proposed to boost the predictive performance. This ensemble model achieved a mIoU of 64.13\% and ranked first in the competition. Notably, our proposed model with a mIoU of 64.52\% outperforms this previous best result, establishing a new state-of-the-art for semantic segmentation on the FLAIR dataset.

\section{Conclusion}
This work investigated the effectiveness of late fusion for semantic segmentation of remote sensing imagery, leveraging the complementary strengths of Very High Resolution (VHR) aerial imagery and Satellite Image Time Series (SITS) data. We proposed a Late Fusion Deep Learning Model (LF-DLM) that integrates a UNetFormer branch for capturing spatial details from aerial imagery and a U-TAE branch for capturing spatio-temporal dynamics from SITS data. The LF-DLM achieved state-of-the-art performance on the FLAIR dataset, a large-scale benchmark for land cover segmentation using multi-source optical imagery. Compared to the UNetFormer model alone, the LF-DLM achieved an improved mIoU of 0.29\%. This signifies the effectiveness of late fusion in combining information from both data sources to enhance segmentation accuracy across various land cover types.

Furthermore, our best model configuration with a mIoU of 64.52\% surpasses the previous state-of-the-art on the FLAIR dataset, demonstrating its robustness and efficiency while adhering to the challenge's inference time constraints. These findings highlight the potential of late fusion deep learning models for improving the accuracy and robustness of semantic segmentation in remote sensing applications. Future work can explore label-specific fusion techniques and class imbalance mitigation strategies to address remaining challenges and further enhance performance for specific land cover types.

\bibliographystyle{unsrt}  
\bibliography{references}  

\end{document}